# Fundus to Fluorescein Angiography Video Generation as a Retinal Generative Foundation Model

**Running title: Fundus2Video as a Foundation Model**

**Authors**

Weiyi Zhang, MS[1], Jiancheng Yang, PhD[2], Ruoyu Chen, MD[1], Siyu Huang, PhD[3], Pusheng Xu, MD[1], Xiaolan Chen, MD[1], Shanfu Lu, MS[1], Hongyu Cao, MS[4], Mingguang He, MD, PhD[1,5,6] [*], Danli Shi, MD, PhD[1,5*]

**Affiliations**

1. School of Optometry, The Hong Kong Polytechnic University, Kowloon, Hong Kong SAR, China.
2. Swiss Federal Institute of Technology Lausanne (EPFL), Lausanne, Switzerland.
3. School of Computing, Clemson University, Clemson, SC, USA.
4. University of Science and Technology of China, Anhui, China.
5. Research Centre for SHARP Vision (RSCV), The Hong Kong Polytechnic University, Kowloon, Hong Kong SAR, China.
6. Centre for Eye and Vision Research (CEVR), 17W Hong Kong Science Park, Hong Kong SAR, China.

**Correspondence**

**Dr. Danli Shi,** MD, PhD., Research Assistant Professor, School of Optometry, The Hong Kong Polytechnic University, Kowloon, Hong Kong SAR, China.

Email: danli.shi@polyu.edu.hk

**Prof. Mingguang He,** MD, PhD., Chair Professor of Experimental Ophthalmology, School of Optometry, The Hong Kong Polytechnic University, Kowloon, Hong Kong SAR, China.

Email: mingguang.he@polyu.edu.hk

## Abstract

Fundus fluorescein angiography (FFA) is crucial for diagnosing and monitoring retinal vascular issues but is limited by its invasive nature and restricted accessibility compared to color fundus (CF) imaging. Existing methods that convert CF images to FFA are confined to static image generation, missing the dynamic lesional changes. We introduce Fundus2Video, an autoregressive generative adversarial network (GAN) model that generates dynamic FFA videos from single CF images. Fundus2Video excels in video generation, achieving an FVD of 1497.12 and a PSNR of 11.77. Clinical experts have validated the fidelity of the generated videos. Additionally, the model's generator demonstrates remarkable downstream transferability across ten external public datasets, including blood vessel segmentation, retinal disease diagnosis, systemic disease prediction, and multimodal retrieval, showcasing impressive zero-shot and few-shot capabilities. These findings position Fundus2Video as a powerful, non-invasive alternative to FFA exams and a versatile retinal generative foundation model that captures both static and temporal retinal features, enabling the representation of complex inter-modality relationships.

**Keywords:** Video generation, Generative adversarial network, Foundation model, Representation, Fluorescein angiography

# Introduction

Color fundus photography (CF) and fundus fluorescein angiography (FFA) are vital diagnostic tools for retinal diseases. CF is favored for its simplicity and affordability, particularly in resource-limited settings like rural and remote areas. However, CF has limitations in detecting lesions like macular edema, retinal non-perfusion, and neovascularization, essential for diagnosing vascular diseases like diabetic retinopathy (DR) and guiding treatment. In contrast, FFA excels in identifying and monitoring these lesions by using an injected dye to dynamically highlight pathological changes. Despite its diagnostic benefits, FFA is invasive and carries risks of potential side effects, including nausea and anaphylactic shock.[1] Therefore, developing non-invasive, safe, and cost-effective alternatives is crucial.

Generative adversarial networks (GANs) have shown promise in translating CF images into realistic FFA images, as demonstrated by the Fundus2FA model.[2,3] This method can robustly translate CF into realistic FFA images and improve DR stratification. However, existing methods are limited to generating static snapshots, lacking the continuous and dynamic vascular changes observed in real FFA.[4] We, therefore, introduced Fundus2Video[5], an autoregressive GAN capable of generating video sequences from CF. However, this pilot study was constrained by a small dataset size and did not fully evaluate the clinical value of the generated videos. In this study, we address these limitations and thoroughly assess the clinical utility of Fundus2Video.

Recent advancements in deep learning have led to the development of foundational models capable of performing general tasks.[6] These models are pre-trained on large datasets to extract generalized feature representations, which can be easily adapted to specific tasks by fine-tuning with small amounts of labeled data. In ophthalmology, foundational models based on the masked autoencoder[7,8] and the CLIP framework[9,10] have emerged. Despite their success, these models require large datasets, extensive training time, and significant computational resources. Additionally, they are mainly tested in classification tasks. To our knowledge, no work demonstrates that a lightweight cross-modality model can generalize to a diverse range of downstream tasks, including segmentation and multiple classification.

To address the challenges inherent in dynamic FFA generation and to harness the potential of generative models as domain-specific foundation models, we propose Fundus2Video, a novel framework that generates FFA videos from static CF images. This cross-modality generative model excels at video generation and serves as a universal pretrained model for various tasks, including ophthalmic and systemic disease classification, segmentation, and image retrieval across ten multinational datasets, under zero-shot, few-shot, and supervised experimental settings. Fundus2Video reveals the transformative potential of lightweight modality translation frameworks, positioning them as next-generation generative foundation models.

# Results

Fig. 1 illustrates the study design. Fundus2Video was developed using 1,956 CF images paired with their corresponding FFA videos. These videos encompass frames from the entire FFA examination process, including the arterial, venous, and late phases, totaling 72,851 FFA frames. The dataset includes various lesions such as macular edema, microaneurysms, capillary leakage, and retinal non-perfusion. Fundus2Video employs an autoregressive GAN architecture, enhanced by an unsupervised clinical knowledge mask (Fig. 1a). To demonstrate its versatility as a retinal foundation model, we evaluated Fundus2Video across ten distinct datasets, covering a wide range of tasks (Fig. 1b).

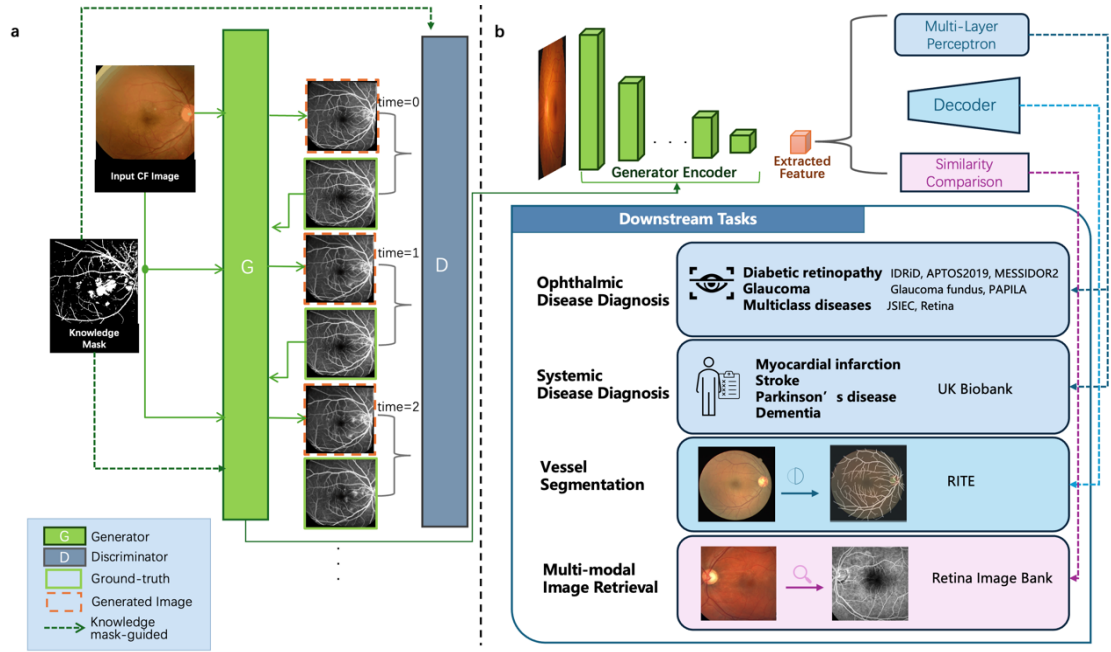

**Fig. 1:** Study overview. (a) Architecture of the Fundus2Video Model: The model is based on an autoregressive high-resolution GAN. The generator produces one frame at a time, using the output from the previous time step and the CF image embedded in different channels as input. During training, the network is guided by an unsupervised knowledge mask, created by calculating the difference between the first and last frames. This approach leverages clinical insights that more severe lesions result in more significant fluorescent changes. The knowledge mask effectively highlights pathological regions in the FFA frames. (b) Adapting Fundus2Video for Various Downstream Tasks: the generator encoder from Fundus2Video was used to accomplish various downstream tasks, including ophthalmic disease diagnosis, systemic disease prediction, vessel segmentation, and multi-modal image retrieval.

## Quality of Generated FFA Videos

**Model comparisons.** We evaluated Fundus2Video against existing image-to-video translation methods, including the auto-encoder-based Seg2Vid[11], diffusion model-based Med-ddpm[12] and ConsistI2V[13]. Each model generates a 12-frame video covering the entire FFA examination process, including the vascular, venous, and late phases. Table 1 shows that our model outperforms all others across various metrics, achieving the lowest Fréchet Video Distance (FVD) of 497.12 and Learned Perceptual Image Patch Similarity (LPIPS) of 0.2131, along with the highest Structural Similarity Index (SSIM) of 0.3795 and Peak Signal-to-Noise Ratio (PSNR) of 11.77.

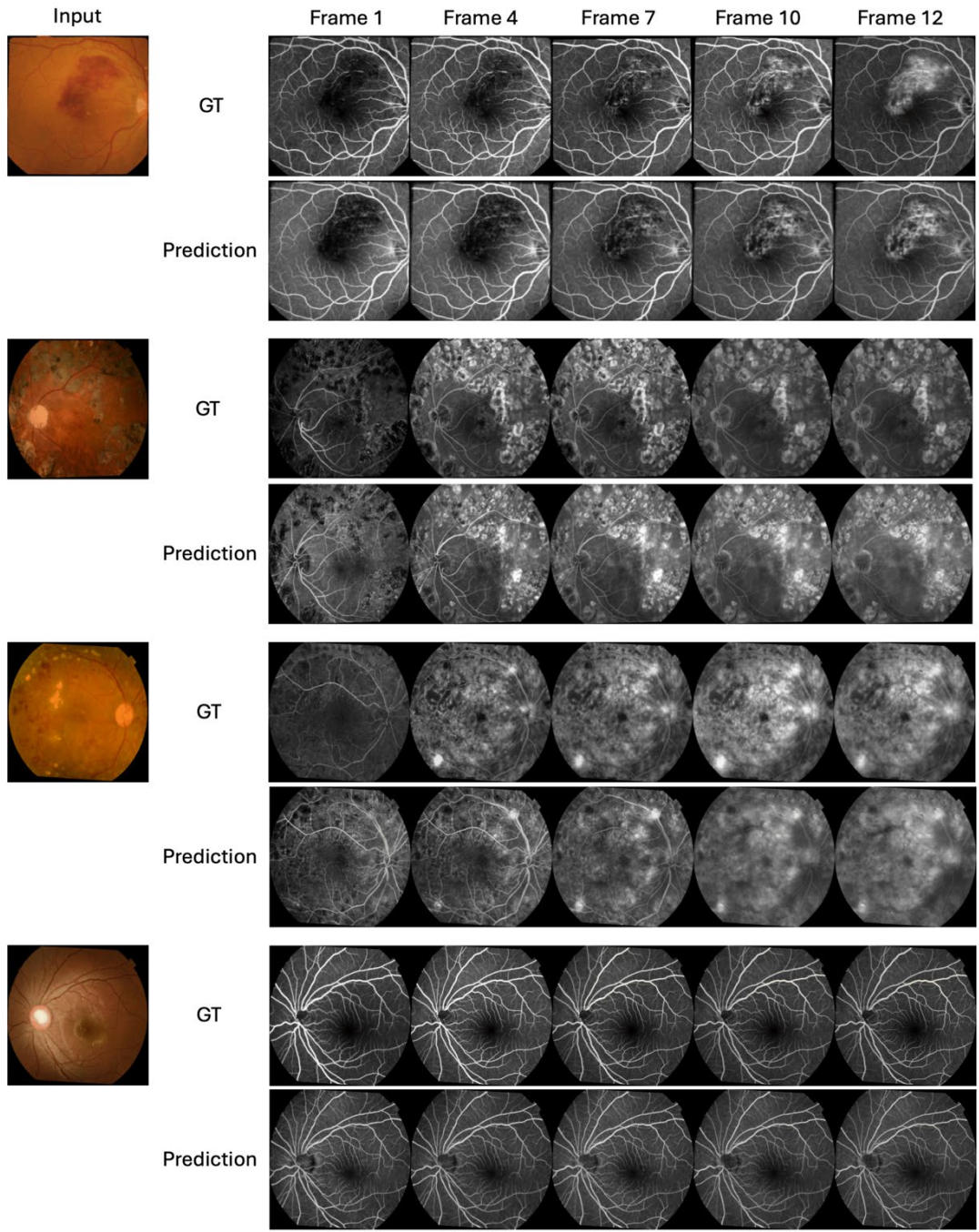

**Fig. 2:** Visualization examples of generated videos. GT = Ground Truth. The original generated videos consist of 12 frames, with 5 frames sampled for display. The first case includes lesions of leakage and nonperfusion areas. The second and third cases feature microaneurysms, leakage, nonperfusion areas, and laser scars. The fourth case shows no abnormalities in FFA. Fundus2Video effectively generates smooth FFA series that accurately replicate anatomical and pathological features.

**Table 1.** Comparison of video generation methods. Fundus2Video outperforms other models across various metrics, including FVD, SSIM, PSNR, and LPIPS. It demonstrates superior performance over the auto-encoder-based Seg2Vid, as well as the diffusion model-based Med-ddpm and ConsistI2V.

| Models | FVD↓ | SSIM↑ | PSNR↑ | LPIPS↓ |
|---|---|---|---|---|
| **Seg2Vid** | 2301.23 | 0.2871 | 10.54 | 0.2513 |
| **Med-ddpm** | 2401.43 | 0.2552 | 10.73 | 0.2407 |
| **ConsistI2V** | 2231.35 | 0.2598 | 10.84 | 0.2399 |
| **Fundus2Video** | **1497.12** | **0.3795** | **11.77** | **0.2131** |

**Visualization results.** Visualization examples are shown in Fig. 2. Fundus2Video produces realistic FFA videos regarding retinal structures and lesional areas, such as microaneurysms, leakage, nonperfusion area, laser scar, etc.

**Human assessment.** Two ophthalmologists reviewed the results from all methods in Table 1 and concluded that Fundus2Video significantly outperformed the others. They then conducted a quality assessment on 50 randomly selected FFA videos generated by Fundus2Video from the test set. The evaluation focused on vascular perfusion, lesion dynamics, and overall coherence, comparing the generated videos to corresponding CF images and ground-truth FFA videos. Scores ranged from 1 (excellent quality) to 5 (very poor quality). The detailed rules and results are presented in Table 2. The experts gave subjective quality scores of 1.28 and 1.34, respectively, with substantial agreement and no statistical difference indicated by a kappa value of 0.90 and a P-value less than 0.001. This high level of agreement underscores the reliability and consistency of Fundus2Video's video generation quality.

**Table 2.** Human visual assessment results. A. Results. B. Assessment criteria. The two aspects of the scoring criteria are vascular perfusion and conformity of lesion dynamics to clinical patterns in the video. Scores ranged from 1 (excellent quality) to 5 (very poor quality).

| A. Results. | | | |
|---|---|---|---|
| Score by Rater 1 | Score by Rater 2 | Kappa value | P-value (95% confidence level) |
| 1.28 | 1.34 | 0.90 | <0.001 |
| B. Assessment criteria. | | | |
| Scale | Quality | Description | |

| | | |
|---|---|---|
| 1 | Excellent | Similar morphology and fluorescent changes of retinal vessels and lesions |
| 2 | Good | Slight morphology and fluorescent changes of retinal vessels and lesions |
| 3 | Normal | Moderate variation in morphology and fluorescent changes of retinal vessels and lesions |
| | | Indicates lesions of diagnostic significance |
| 4 | Poor | Large variation in morphology and intensity of retinal vessels and lesions |
| | | Without lesions of diagnostic significance |
| 5 | Very Poor | Large variation in morphology and intensity of choroidal and retinal vessels |

## Downstream Performance in Zero-shot and Few-shot Setting for Vessel Segmentation

Fundus2Video's generator exhibits strong zero-shot and few-shot segmentation capabilities. We validated this on the vascular segmentation dataset RITE[14], conducting few-shot experiments on retinal vessels with support numbers of 0, 1, 2, 4, 8, and 16. We compared Fundus2Video with Fundus2FA, a GAN model based on Pix2PixHD[15] generating venous and late-phase single frame FFA (here we used the venous-phase model weight)[16], and the recent segmentation foundation model Medical-SAM2[17], known for its medical generalization capability. Detailed numerical comparisons and visual results are shown in Fig. 3.

As illustrated in the bar charts of Fig. 3(a), Fundus2Video outperformed all models in all settings in terms of the area under the receiver operator characteristic curve (AUROC) and Dice scores, particularly compared to Medical-SAM2 and Pix2pixHD. Notably, Fundus2Video achieves an AUROC of 0.943 in the zero-shot setting and a Dice score of 0.733 in the one-shot setting, showcasing its powerful few-shot segmentation capabilities. The visual comparison of one-shot segmentation results in Fig. 3(b) further emphasizes Fundus2Video's superiority, where other methods display some loss of detail in vascular segmentation, while Fundus2Video's results closely resemble the ground truth with remarkably rich details. This demonstrates Fundus2Video's profound understanding and cross-modal reconstruction ability for fundus images, particularly in capturing fine-grained vascular structures.

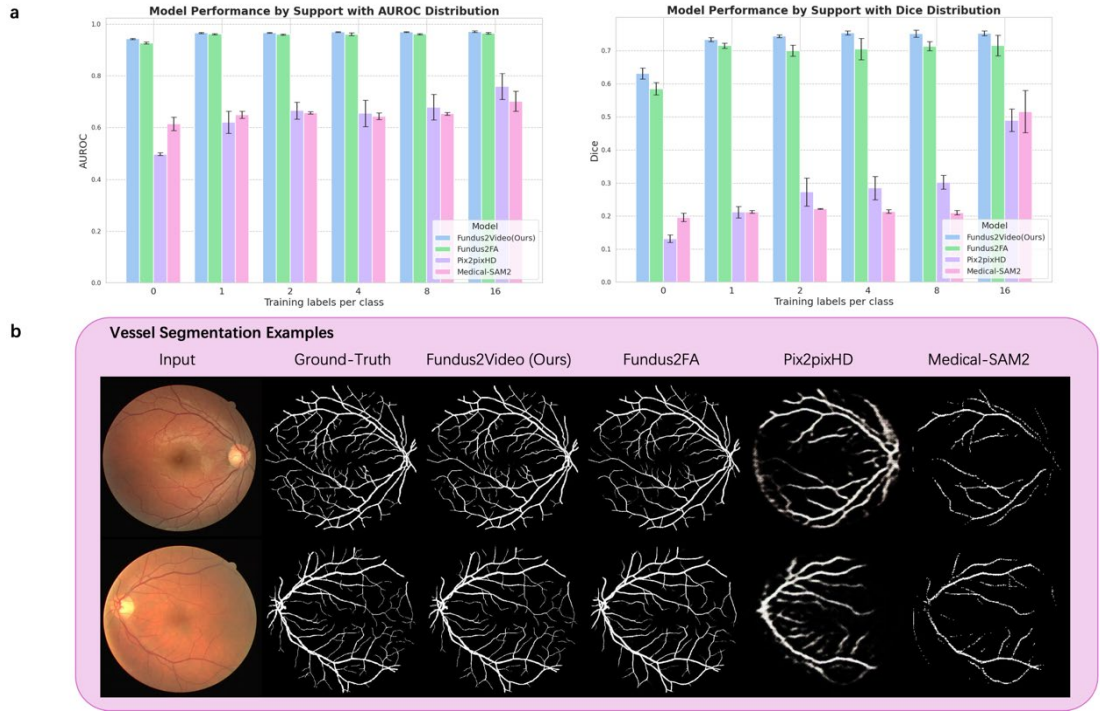

**Fig. 3:** Vessel segmentation results. (a) Model performance comparison. AUROC = area under the receiver operator characteristic curve. We compared the segmentation capabilities of four models by setting the number of training samples to 0, 1, 2, 4, 8, and 16. Five experiments were conducted for each setting with different datasets randomly sampled and divided. Fundus2Video achieved the best AUROC and Dice scores across all settings, significantly outperforming Pix2pixHD and Medical-SAM2. (b) Vessel segmentation visualization results of the one-shot setting. The output of Fundus2Video closely approximates the ground truth with rich details, whereas the results from other models are visibly incomplete.

## Downstream Performance in Zero-shot, Few-shot, and Supervised Training for Ophthalmic Disease Diagnosis

To evaluate Fundus2Video's potential as a foundation model for diagnosing various ophthalmic diseases from CF images, we evaluated it on seven public datasets. We compared Fundus2Video with Fundus2FA and RETFound, the previous state-of-the-art foundation model in ophthalmology. All results were averaged over five runs with different random seeds.

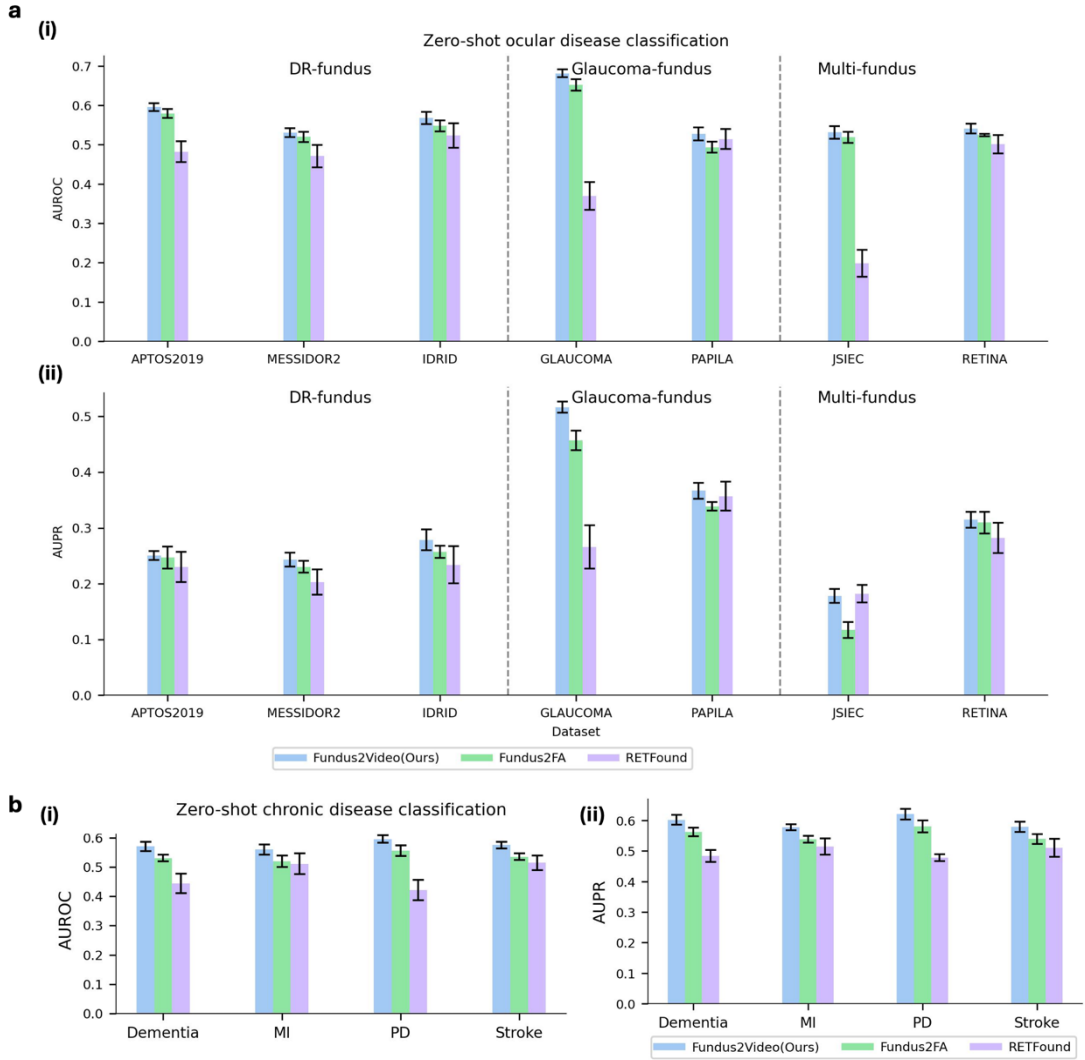

**Fig. 4:** Zero-shot disease classification results. AUROC = area under the receiver operator characteristic curve, AUPR = area under the precision-recall curve, MI = Myocardial Infarction, PD = Parkinson's Disease. (a) Results of ocular disease classification. Fig a.(i) shows the AUROC results, and Fig a.(ii) shows the AUPR results. Experiments on each dataset were averaged over five runs. Fundus2Video achieved a higher mean AUROC compared to the other two models, Fundus2FA and RETFound, across all datasets (P < 0.001). The P-value was calculated by comparing the prediction results of Fundus2Video with the prediction from the other model with a higher AUROC result using a t-distribution. (b) Results of chronic disease classification. Fig b.(i) shows the AUROC results, and Fig b.(ii) shows the AUPR results. Experiments on each dataset were averaged over five runs. Fundus2Video demonstrated significantly better AUROC and AUPR across all tasks, with all P-values < 0.001.

In the zero-shot classification, where the encoder from the models was used without further training, Fundus2Video consistently outperformed others across all datasets. Notably, on glaucoma datasets such as Glaucoma Fundus[18] and the multi-disease dataset JSIEC[19],

Fundus2Video achieved the area under the receiver operator characteristic curve (AUROC) scores more than 30 percent higher than RETFound. This demonstrates Fundus2Video's strong ability to understand features from external fundus images across diverse sources. Results are presented in Fig. 4(a) and Supplementary Table 1.

We further evaluated the models' performance in few-shot learning settings with 1, 2, 4, and 8 support examples per class. Fundus2Video showed a clear advantage in the one-shot setting across all seven datasets, indicating strong feature generalization and learning capabilities even with minimal labeled data. The results suggest that Fundus2Video effectively captures essential retinal features with limited supervision. Detailed results are shown in Fig. 5(a) and Supplementary Table 2.

Finally, we fine-tuned and tested the models by splitting the datasets into 55:15:30% (train:val: test). The corresponding results are presented in Fig. 6(a) and Supplementary Table 3. In diabetic retinopathy datasets (MESSIDOR-2, IDRID[20], and APTOS-2019), Fundus2Video improved AUROC by up to 1.5% (e.g., MESSIDOR-2) and increased area under the precision-recall curve (AUPR) scores, demonstrating better classification of disease stages compared to using static CF images alone. For glaucoma-focused datasets (PAPILA[21] and Glaucoma Fundus[18]), Fundus2Video achieved AUROC improvements of up to 0.8% and maintained high AUPR values, indicating enhanced sensitivity and specificity in detecting glaucoma-related changes. Across multi-class eye disease datasets (JSIEC[19] and Retina), Fundus2Video consistently raised AUROC by up to 0.6% and improved AUPR, with statistically significant improvements (p-value < 0.001). These findings underscore Fundus2Video's efficacy in enhancing overall classification accuracy across diverse eye conditions, emphasizing the utility of dynamic FFA data in improving diagnostic precision.

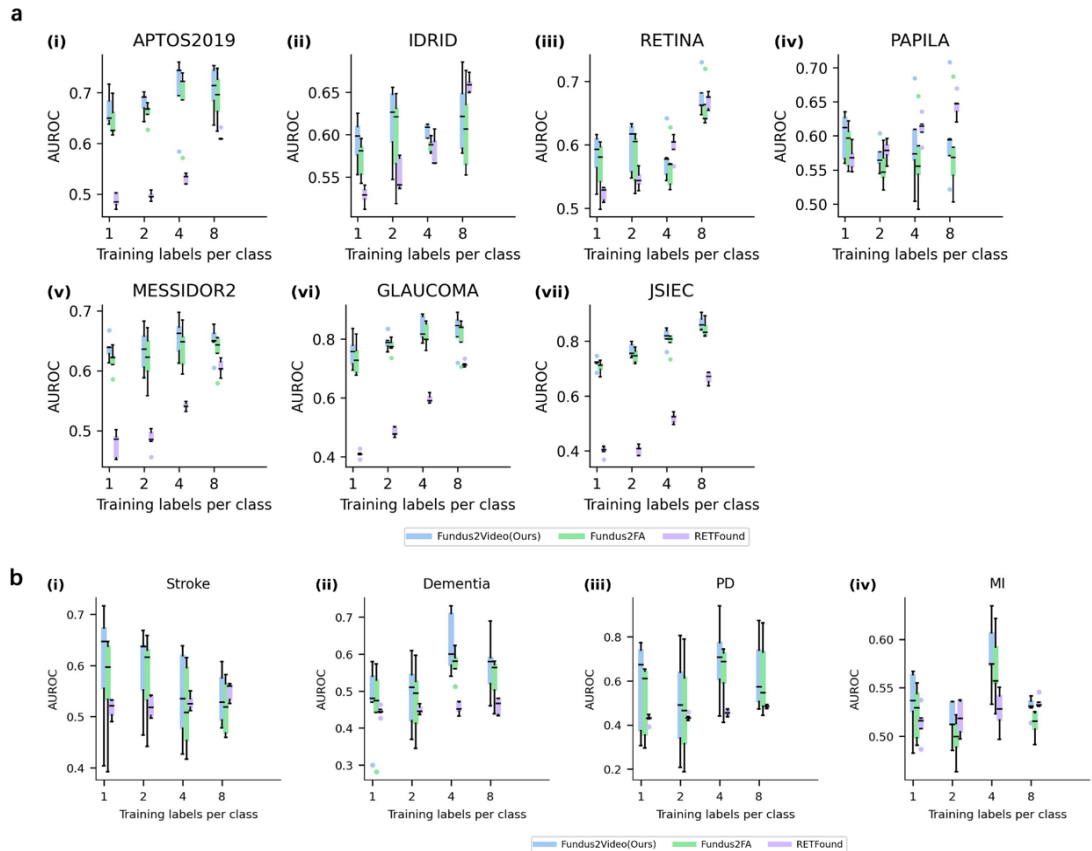

**Fig. 5:** Few-shot classification results. AUROC = area under the receiver operator characteristic curve. We compared the few-shot learning capabilities of different models across various datasets, considering multiple support numbers for each class. For each support number, we sampled five different sets of training examples and conducted five times of experiments. The boxes represent the quartiles, and the whiskers extend to data points within 1.5 times the interquartile range. AUPR results are shown in Supplementary Fig.1. (a) Ophthalmic diseases. We selected support numbers of 1, 2, 4, and 8. Notably, when the support number is 1, Fundus2Video significantly outperforms the other two models across all datasets (P<0.001), demonstrating its exceptional generalization capability. Additionally, on the APTOS2019, MESSIDOR2, Glaucoma, and JSIEC datasets, Fundus2Video consistently outperforms the other two models across all support numbers (P<0.001). (b) Chronic diseases. We selected support numbers of 1, 2, 4, 8, and 16. In the few-shot results for chronic systemic diseases, Fundus2Video achieves superior performance compared to the other two models across all support numbers for all diseases except stroke (P<0.001). For stroke, Fundus2Video also shows excellent performance with support numbers of 1, 2, and 4.

## Downstream Performance in Zero-shot, Few-shot, and Supervised Training for Systemic Disease Diagnosis

To evaluate Fundus2Video's improvement in systemic disease diagnosis using CF images, we compared it with Fundus2FA and RETFound across Stroke, Dementia, Parkinson's Disease (PD), and Myocardial Infarction (MI) using the UK Biobank dataset[14]. Similar to our ophthalmic disease tests, we evaluated the models in zero-shot, few-shot (support number=1, 2, 4, 8, 16), and supervised learning settings. Results are shown in Fig. 3-5 (b) and Supplementary Table 4-6. Fundus2Video consistently achieved the highest AUROC and AUPR scores, with statistically significant improvements for all four diseases compared to Fundus2FA and RETFound (all $p < 0.001$).

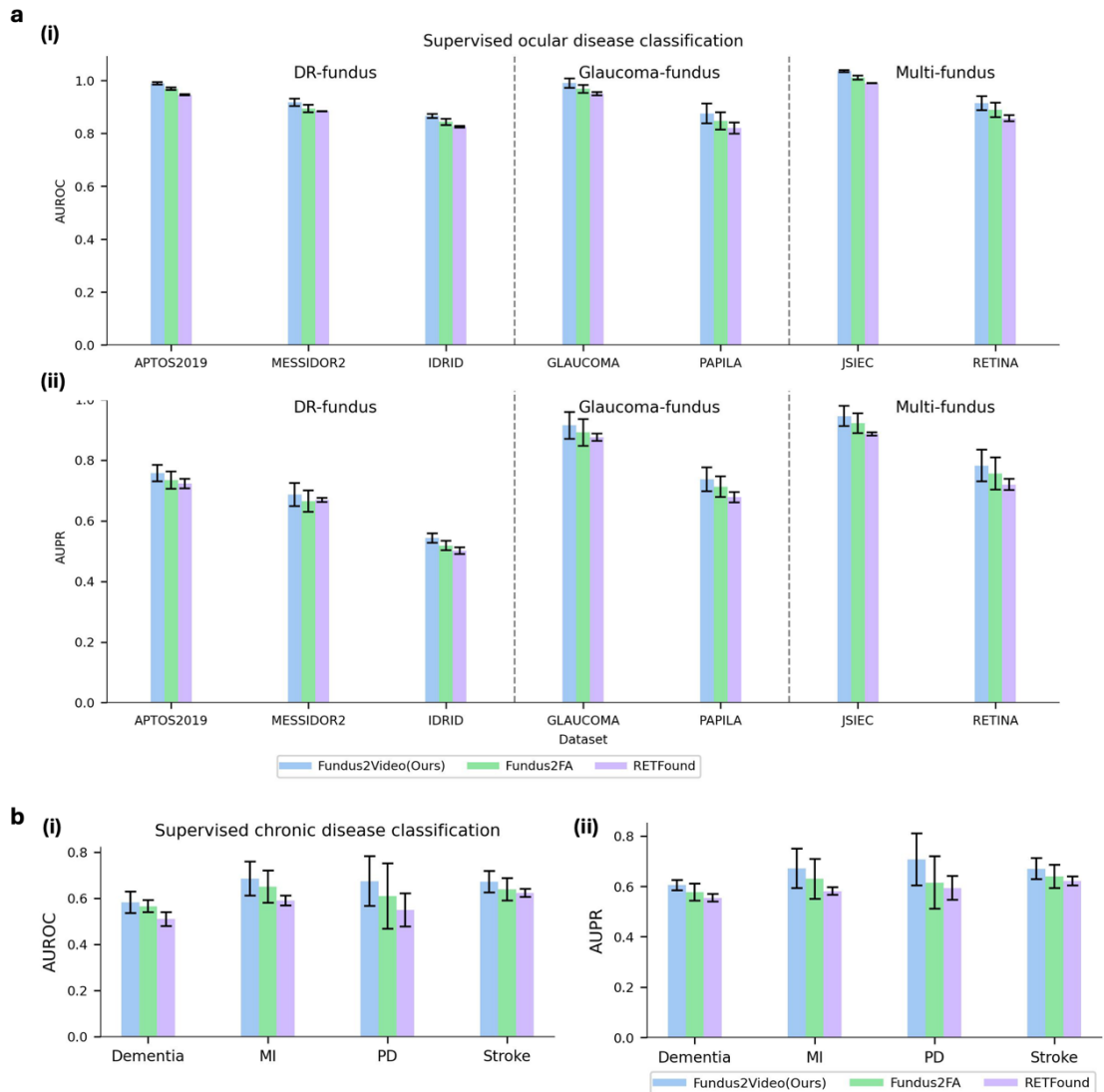

**Fig. 6:** (a) Results of ocular disease classification. Fig a.(i) presents the AUROC results, while Fig a.(ii) displays the AUPR results. AUROC = area under the receiver operator characteristic curve, AUPR = area under the precision-recall curve. Each dataset's experiments were averaged over five runs. Fundus2Video achieved a higher mean AUROC compared to the other two models, Fundus2FA and RETFound, across all datasets. Regarding AUPR, Fundus2Video outperformed the other two models except for the MESSIDOR2 dataset. (b) Results of chronic

disease classification. Fig b.(i) shows the AUROC results, and Fig b.(ii) shows the AUPR results. Each dataset's experiments were averaged over five runs. For AUROC, Fundus2Video was the best performer across all tasks except for Dementia, which was slightly lower than Fundus2FA (P>0.05). Fundus2Video also demonstrated superior AUPR across all tasks.

Notably, in the zero-shot setting, Fundus2Video outperformed RETFound by over ten percentage points in both AUROC and AUPR when predicting Dementia and PD. In the one-shot setting, Fundus2Video demonstrated an even greater advantage, exceeding RETFound by 30 percentage points in AUROC and AUPR for PD. This underscores Fundus2Video's exceptional ability to capture retinal features related to vascular changes linked to chronic systemic diseases.

**Downstream Performance in Multi-modal Image Retrieval**

To further explore Fundus2Video's robust feature representation beyond CF and FFA, we investigated its performance in image retrieval across multiple modalities. Accurate image retrieval in clinical practice aids in case comparison, disease progression tracking, and identifying rare conditions, especially with multimodal imaging data. This is particularly useful for searching similar cases in large-scale medical databases or retrieving specific imaging patterns for diagnostic support.

We compared the multimodal image retrieval abilities of Fundus2Video, Fundus2FA, and RETFound. Validation was performed on two external multimodal datasets: AngioReport[22] and Retina Image Bank[23]. The AngioReport dataset includes images from indocyanine green angiography (ICGA) and FFA, while Retina Image Bank encompasses 14 ophthalmic imaging modalities, including rare diseases. Retrieval performance and visualized examples are shown in Fig. 7 and Supplementary Table 7.

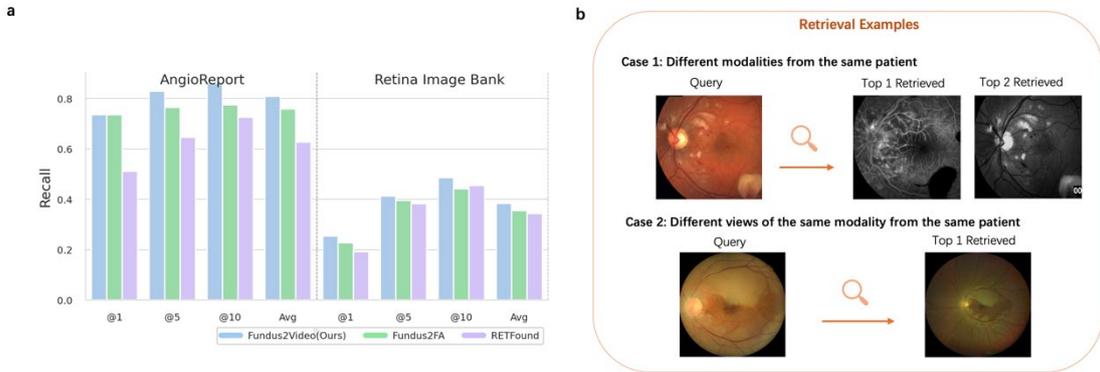

**Fig. 7:** Zero-shot multi-modal image retrieval results. (a) Comparison results of different models. Similarity in the embedding space was computed between the query image and all the other images in the test set. The top K most similar images were retrieved, where K = 1, 5, 10.

We report Recall@K and the mean recall. Fundus2Video outperforms all other models across both datasets. (b) Retrieval examples using Fundus2Video. Case 1 demonstrates the model's ability to retrieve images of different modalities from the same patient, while Case 2 shows the model's capability to retrieve images of different views within the same modality from the same patient.

Across both datasets, Fundus2Video achieved the best performance in terms of top-1, top-5, top-10, and average recall. In the AngioReport dataset, Fundus2Video's average recall exceeded 0.8. As shown in the retrieval examples in Fig. 7(b), Fundus2Video not only successfully performed cross-modality retrieval for the same patient (Case 1), but also enabled retrieval of images from the same modality and patient with different viewing angles (Case 2). These results highlight Fundus2Video's exceptional ability to handle multimodal images.

## Discussion

Multi-modality examinations provide complementary information, and we bridge this gap by generating non-invasive FFA videos from static CF images. Beyond visual authenticity validated by retinal specialists, we demonstrated its potential as a retinal foundation model across multiple downstream tasks, including vessel segmentation, retinal disease diagnosis, systemic disease prediction, and multimodal retrieval. Fundus2Video captures both cross-modal static and temporal retinal features, enabling the representation of complex inter-modality relationships.

A strength of Fundus2Video is its strong performance in classifying ophthalmic diseases like DR and glaucoma. DR involves microvascular changes such as capillary damage and neovascularization, best viewed by FFA.[24] While glaucoma causes optic nerve damage and progresses silently until significant vision loss occurs.[25] Fundus2Video enhances detection by learning structures like retinal vessels, optic discs, and lesions that are evident on FFA, surpassing static fundus image representations.[7] Its ability to generalize across diverse datasets makes it a powerful tool for large-scale screening, particularly for aging and diabetic populations where accurate and efficient diagnosis is critical.

Moreover, Fundus2Video excels in retinal vessel segmentation, particularly in few-shot and zero-shot learning settings. Fundus2Video achieves near ground-truth accuracy with only one training sample(AUROC=0.965). This indicates that the model has effectively learned retinal microvessels prominent on FFA and is generalizable, which aligns with our previous study using cross-modal pretrain to facilitate vessel segmentation.[26] The fine-grained vascular information learned by Fundus2Video may potentially help in diagnosing both ophthalmic and systemic diseases, as evident in the improvement in ocular and systemic disease prediction, which is a key limitation in many transformer-based models like RETFound that struggle with local feature representation.

Another contribution of Fundus2Video is to demonstrate that generative models can be lightweight, task-specific foundation models. Unlike traditional, larger models, Fundus2Video efficiently learns from CF and FFA modalities with fewer computational demands. Its generator encoder has about 60 million parameters, compared to the 307 million in Vision Transformer Large (ViT-Large) used in RETFound.[7,27] This efficiency enables high performance across diverse tasks, including segmentation, without the burdens of larger architectures.

Discriminative and generative foundation models each have unique strengths.[28] Discriminative models are ideal for classification tasks and perform well with large labeled datasets, valued for their simplicity and effectiveness. On the other hand, generative models are used for self-reconstruction in self-supervised learning or creating synthetic datasets for data augmentation, which is beneficial in the medical field where labeling is costly and time-consuming. Fundus2Video bridges these approaches by integrating both methodologies. It maximizes clinical data utility without extensive labeling, achieving robust performance across multiple tasks and highlighting the potential of generative models to tackle critical challenges in medical imaging while retaining the advantages of discriminative methods.

While Fundus2Video delivers promising results, there are limitations. First, the generated FFA needs further validation in clinical settings. Additionally, except for FFA, incorporating other modalities and multi-faceted information has potent potential to improve the model's capability in downstream tasks, warranting further research.[29,30] The current framework is designed for pixel-aligned modalities; for those that are not, methods like contrastive learning may be more suitable.

In conclusion, this study presents Fundus2Video, an innovative cross-modality generative model that transforms static CF images into dynamic FFA videos while excelling in feature representation for downstream tasks. By learning from both CF and FFA modalities, Fundus2Video enhances diagnostic accuracy for various retinal diseases and shows significant potential for systemic disease screening. Although further clinical validation is needed, Fundus2Video demonstrates transformative potential in ophthalmology and beyond.

## Methods

### Study Population

**Data for algorithm development.** The dataset used for developing Fundus2Video comprises 1,956 CF images and 72,851 corresponding FFA frames obtained from 1,360 anonymized patients who underwent FFA examinations between 2016 and 2019 in China.[2] The CF images were captured using Topcon TRC-50XF and Zeiss FF450 Plus cameras, with resolutions ranging from 1,110 × 1,467 to 2,600 × 3,200 pixels. Meanwhile, FFA images were acquired using Zeiss FF450 Plus and Heidelberg Spectralis cameras, featuring a resolution of 768 × 768 pixels. All patient data were anonymized and deidentified in compliance with Institutional Review Board approvals. During training, the ground-truth FFA videos also consist of 12 frames, with four frames uniformly selected from each arterial, venous, and late phase.

**Data for downstream tasks.** For downstream tasks related to ophthalmic disease diagnosis, we utilized seven publicly datasets. Specifically, Kaggle APTOS-2019 (India), IDRID (India)[20], and MESSIDOR-2 (France) were employed for diabetic retinopathy (DR) diagnosis, graded according to the International Clinical Diabetic Retinopathy Severity Scale encompassing five stages from no DR to proliferative DR. PAPILA (Spain)[21] and Glaucoma Fundus (South Korea)[18] datasets focused on glaucoma classification, spanning non-glaucoma, early glaucoma (suspected), and advanced glaucoma stages. For multi-class eye diseases, we utilized the JSIEC (China)[19] and Retina datasets. JSIEC comprises 1,000 images covering 39 fundus diseases, while the Retina dataset includes labels for normal, glaucoma, cataract, and retinal diseases. Supplementary Table 1 presents the characteristics of the datasets.

For systemic disease prediction, we used the UK Biobank dataset[14] focusing on predicting cardiovascular and neurological diseases over a 5-year period. This dataset includes 502,665 UK residents aged 40-69 years registered with the National Health Service, among whom 82,885 underwent comprehensive fundus photography (3D OCT-1000 Mark II, Topcon, Japan), resulting in 171,500 CF images. Systemic diseases evaluated include stroke, dementia, Parkinson's disease (PD), and myocardial infarction (MI). For each disease, we performed balanced sampling between the disease and healthy groups across different age ranges. We used only the CF images of the right eye captured during a single visit per patient as input for downstream tasks to ensure consistency and minimize biases.

For the downstream retinal vessel segmentation, we used the RITE[31], which consists of 40 retinal fundus images with corresponding vessel labels. This dataset is widely used for evaluating retinal vessel segmentation algorithms.

For multi-modal image retrieval, we utilized two datasets: AngioReport[22] and Retina Image Bank[23]. The AngioReport dataset contains over 50,000 FFA and ICGA images collected from routine clinics in Thailand. We randomly selected a subset of patients from this dataset, including 387 images for our experiments. The Retina Image Bank is a large public repository from the United States, consisting of images representing 14 modalities and 84 ophthalmic diseases. We focused on images collected between 2019 and 2023, creating a custom dictionary to filter cases only involving rare diseases such as serpiginous choroiditis[32], Stargardt disease[33], retinoblastoma[34] and so on. After this selection, we identified 405 images from 150 patients for our image retrieval tasks. This subset was used to assess the model's ability to retrieve multi-modal and rare-disease cases.

## Model Development

**Overview.** Fundus2Video aims to generate a realistic FFA video $\hat{Y}$ from a given CF image $x$, with the ground-truth FFA video during training represented as $Y$. Considering the temporal nature of the FFA series, we adopt an autoregressive GAN architecture based on pix2pixHD[16], to sequentially generate each frame of the FFA video by incorporating the CF image and preceding frames. Multi-frame input and smoothing techniques ensure temporal consistency, where three consecutive frames from the ground-truth FFA series are input in a sliding window fashion, followed by triple-frame averaging to smooth transitions between frames.

**Knowledge mask-guided video generation.** Since the baseline model may miss fine details in critical regions, we introduce a binary knowledge mask $m$, calculated as the difference between the first and last frames of the FFA video Y:

$$m = \delta(Y_0 - Y_T)$$

where $\delta$ is a threshold function. This mask helps the model focus on areas of importance, improving the depiction of lesions and key structures.

We integrate the knowledge mask into the model through several techniques:

Knowledge-boosted Attention: The model's attention is guided toward focal regions using an attention-loss $\mathcal{L}_{Att}$, which compares the attention map $A$ to the knowledge mask $m$:

$$\mathcal{L}_{Att} = \frac{1}{n}\sum(A_i - m_i)^2$$

Mask-enhanced Losses $\mathcal{L}_{Mask}$: We incorporate the knowledge mask into the PatchNCE loss[35] to encourage the network to prioritize informative regions during training.

Knowledge-aware Discriminators: Multiple discriminators assess the images at different scales, using the mask to focus on relevant regions.

The final objective function $\mathcal{L}$ combines these techniques:

$$\mathcal{L} = \lambda_{Mask}\mathcal{L}_{Mask} + \lambda_{Att}\mathcal{L}_{Att} + \lambda_{GAN}\mathcal{L}_{GAN}$$

Where $\mathcal{L}_{GAN}$ is the discriminator loss, and $\lambda_{Mask}$, $\lambda_{Att}$ and $\lambda_{GAN}$ represent the weights assigned to different components of the loss function. These weights are set to 1, 4, and 2 to balance feature extraction, attention, and adversarial learning in the training process..

**Evaluation Criteria**

**Objective evaluation criteria of video quality.** Our objective evaluation criteria include Fréchet Video Distance (FVD)[36], Structural Similarity Index (SSIM)[37], Peak Signal-to-Noise Ratio (PSNR)[38], and Learned Perceptual Image Patch Similarity (LPIPS)[39]. FVD measures the similarity of feature distributions between real and generated videos, capturing overall quality and coherence. SSIM evaluates the structural similarity between the generated and ground truth videos, focusing on luminance, contrast, and structural information. PSNR quantifies reconstruction quality by comparing pixel-wise differences, with higher values indicating a closer resemblance to the ground truth. LPIPS assesses perceptual similarity using deep network features to compare image patches, capturing perceptual differences not evident through traditional pixel-based metrics. These metrics collectively ensure a thorough evaluation of the generated FFA videos, covering statistical similarity, structural fidelity, and perceptual quality.

**Subjective evaluation criteria of video quality.** The generated videos were evaluated against real FFA videos on the test set. Two ophthalmic specialists, trained to follow a predefined consensus, assessed the videos by rating the generated videos compared to the real ones in terms of lesion integrity, retinal and choroidal structures, and dynamic range. Detailed evaluation criteria are provided in Table 2.

**Downstream classification evaluation.** In evaluating classification downstream tasks, we employed AUROC and AUPR metrics to assess classification performance using receiver operating characteristics and precision-recall curves, respectively. For binary classification tasks, such as diagnosis of systemic diseases, AUROC and AUPR were computed in a binary context. For multi-class classification tasks, such as five-stage diabetic retinopathy (DR) and multi-class disease diagnoses, AUROC and AUPR were calculated individually for each disease class and then averaged (macro) to derive overall AUROC and AUPR scores.

**Downstream segmentation evaluation.** We evaluated vessel segmentation performance using the AUROC and Dice coefficient. AUROC measures the model's ability to distinguish between true and false positives across different thresholds, while Dice quantifies the overlap between predicted and ground truth segmentations. Higher values for both indicate better segmentation accuracy.

**Downstream retrieval evaluation.** For multi-modal image retrieval, we used Recall@K, where K is 1, 5, and 10, to assess the proportion of correct images retrieved within the top K results. The average recall across these values provides an overall measure of retrieval performance.

**Robustness and statistical evaluation.** To ensure robustness, each task underwent evaluation across 5 random splits, with models trained using different random seeds to mitigate variability. Mean performance across these iterations was computed, and 95% confidence intervals (CI) were derived using 1.96 times the standard error. We determined the statistical significance between the performance of Fundus2Video and the best-performing model among the others using two-sided t-tests.

## Comparison Models

**Pix2pixHD[16]:** Pix2pixHD is a GAN-based model designed for high-resolution image generation tasks, particularly for pixel-to-pixel translation between different modalities. It utilizes a multi-scale generator and discriminator to generate high-quality, realistic images that capture fine details. This model has been widely adopted in various fields, including medical imaging, because it can generate clear and consistent outputs in cross-modality translation tasks.

**Fundus2FA[2]:** Fundus2FA is a specialized model built on pix2pixHD, designed explicitly for generating FFA images from Color Fundus CF. It is optimized for ophthalmology tasks and has been validated on downstream classification tasks related to diabetic retinopathy (DR). The model successfully generates realistic FFA images that capture essential vascular features.

**RETFound[7]:** RETFound is the first foundation model developed for ophthalmology, trained separate weights for CF and Optical Coherence Tomography (OCT). Using 1.6 million unlabeled retinal images for self-supervised learning, RETFound has established itself as a versatile model capable of generalizing across various classification tasks, including disease diagnosis for retinal diseases like DR, age-related macular degeneration (AMD), glaucoma, and systemic diseases.

**Medical-SAM2[17]:** Medical-SAM2 is a state-of-the-art segmentation model designed to handle various medical image segmentation tasks. Medical-SAM2 has been adapted to retinal imaging to precisely segment structures like optic disc. It employs a segmentation transformer-based architecture that excels at learning spatial dependencies and has demonstrated high performance in various medical image segmentation benchmarks. The model is recognized for its generalization capabilities, particularly in few-shot learning scenarios, making it suitable for tasks with limited labeled data.

**Adaptation to downstream tasks**

To evaluate the generalization capabilities of Fundus2Video, we employed its generator encoder as a feature extractor for zero-shot and few-shot classification tasks. A Multi-Layer Perceptron (MLP) was integrated with the feature extractor to perform classification. This methodology facilitates the assessment of the model's capacity to capture generalizable retinal features without extensive fine-tuning. Through comparative analysis with analogous adaptations of Fundus2FA and RETFound, we quantitatively evaluated the efficacy of our cross-modal video generation approach in learning transferable features. The MLP outputs a probability distribution over the disease categories, with the final layer containing neurons corresponding to the number of classes. In supervised training classification tasks, our primary goal is to evaluate whether Fundus2Video can enhance the performance of RETFound. To achieve this, we combined the features from the Fundus2Video generator encoder and the RETFound encoder using soft attention. The integrated features are then fed into an MLP for classification.

As for the parameter settings, for zero-shot classification, none of the models were trained on the downstream data; they were directly tested on the classification task. In the few-shot setup, we trained the models with support sets containing 1, 2, 4, or 8 examples per class. For fully supervised learning, we split the data into training, validation, and test sets with a 55:15:30 ratio. We applied label smoothing to better align the output distribution with the true labels. The training was conducted with a batch size of 16 over 50 epochs. The learning rate was gradually increased from 0 to $5\times10^{-4}$ during the first ten epochs, followed by a cosine annealing schedule that reduced the learning rate from $5\times10^{-4}$ to $1\times10^{-6}$ over the remaining 40 epochs. After each epoch, model performance was evaluated on the validation set, and the model weights with the highest AUROC were saved. These best-performing weights were then used for the final test evaluations.

In the downstream task of vessel segmentation, we fine-tuned the pretrained encoders from Fundus2Video, Fundus2FA, and Pix2pixHD. We applied basic transformations such as random brightness and contrast adjustments, cropping, and flipping for augmentation. For the Medical-SAM2 model, we used the pretrained weights from sam2_hiera_small.pt and followed the official fine-tuning protocol. In all models, the few-shot experiments were run for 100 epochs with consistent image resizing to 512x512 pixels. This ensured a fair comparison across methods.

For retrieval tasks, we used the same approach as zero-shot classification to extract image features. We then retrieved the top K images from the aligned latent space that were closest to a given query image. If the query result and the input image originated from the same patient, we considered the retrieval correct. We selected K values from 1, 5, 10 and reported the average recall by taking the mean of the three Recall@K scores.

## Data availability

The authors do not have the right to redistribute the dataset used for training Fundus2Video. The downstream datasets can be accessed by referring to the original paper.

## Code Availability

All the code is accessible on the following GitHub repository: https://github.com/Michi-3000/Fundus2Video

## Funding

The study was supported by the Start-up Fund for RAPs under the Strategic Hiring Scheme (P0048623) from HKSAR, Global STEM Professorship Scheme (P0046113), and Henry G. Leong Endowed Professorship in Elderly Vision Health. The sponsors or funding organizations had no role in the design or conduct of this research.

## Acknowledgments

We thank the InnoHK HKSAR Government for providing valuable supports.

## Contributions

W.Z., J.Y., S.H., M.H., and D.S. conceived the study idea. W.Z. and D.S. built the model and ran experiments. M.H. provided data and computing facilities. W.Z., J.Y., R.C., S.H., P.S., and D.S. contributed to key data interpretation. W.Z. wrote the manuscript. All authors critically revised the manuscript. D.S. was the project administrator.

## Conflict of interest

A patent has been filed for this innovation (CN 202410360491.4).

**Uncategorized References**

1. Kornblau, I.S. & El-Annan, J.F. Adverse reactions to fluorescein angiography: A comprehensive review of the literature. *Surv Ophthalmol* **64**, 679-693 (2019).
2. Shi, D*., et al.* Translation of Color Fundus Photography into Fluorescein Angiography Using Deep Learning for Enhanced Diabetic Retinopathy Screening. *Ophthalmol Sci* **3**, 100401 (2023).
3. Chen, R*., et al.* Generating Multi-frame Ultrawide-field Fluorescein Angiography from Ultrawide-field Color Imaging Improves Diabetic Retinopathy Stratification. *arXiv preprint arXiv:2408.10636* (2024).
4. You, A., Kim, J.K., Ryu, I.H. & Yoo, T.K. Application of generative adversarial networks (GAN) for ophthalmology image domains: a survey. *Eye and Vision* **9**, 6 (2022).
5. Zhang, W*., et al.* Fundus2Video: Cross-Modal Angiography Video Generation from Static Fundus Photography with Clinical Knowledge Guidance. in *International Conference on Medical Image Computing and Computer-Assisted Intervention* 689-699 (Springer, 2024).
6. Moor, M*., et al.* Foundation models for generalist medical artificial intelligence. *Nature* **616**, 259-265 (2023).
7. Zhou, Y*., et al.* A foundation model for generalizable disease detection from retinal images. *Nature* **622**, 156-163 (2023).
8. Shi, D*., et al.* EyeFound: A Multimodal Generalist Foundation Model for Ophthalmic Imaging. *arXiv preprint arXiv:2405.11338* (2024).
9. Silva-Rodriguez, J., Chakor, H., Kobbi, R., Dolz, J. & Ayed, I.B. A foundation language-image model of the retina (flair): Encoding expert knowledge in text supervision. *arXiv preprint arXiv:2308.07898* (2023).
10. Shi, D*., et al.* EyeCLIP: A visual-language foundation model for multi-modal ophthalmic image analysis.   (2024).
11. Pan, J*., et al.* Video generation from single semantic label map. in *Proceedings of the IEEE/CVF Conference on Computer Vision and Pattern Recognition* 3733-3742 (2019).
12. Dorjsembe, Z., Pao, H.-K., Odonchimed, S. & Xiao, F. Conditional diffusion models for semantic 3d medical image synthesis. *Authorea Preprints* (2023).
13. Ren, W*., et al.* Consisti2v: Enhancing visual consistency for image-to-video generation. *arXiv preprint arXiv:2402.04324* (2024).
14. Chua, S.Y.L*., et al.* Cohort profile: design and methods in the eye and vision consortium of UK Biobank. *BMJ Open* **9**, e025077 (2019).

15. Wang, T.-C., *et al.* High-Resolution Image Synthesis and Semantic Manipulation with Conditional GANs. *2018 IEEE/CVF Conference on Computer Vision and Pattern Recognition*, 8798-8807 (2017).
16. Wang, T.-C., *et al.* High-resolution image synthesis and semantic manipulation with conditional gans. in *Proceedings of the IEEE conference on computer vision and pattern recognition* 8798-8807 (2018).
17. Zhu, J., Qi, Y. & Wu, J. Medical sam 2: Segment medical images as video via segment anything model 2. *arXiv preprint arXiv:2408.00874* (2024).
18. Ahn, J.M., *et al.* A deep learning model for the detection of both advanced and early glaucoma using fundus photography. *PloS one* **13**, e0207982 (2018).
19. Cen, L.-P., *et al.* Automatic detection of 39 fundus diseases and conditions in retinal photographs using deep neural networks. *Nature communications* **12**, 4828 (2021).
20. Porwal, P., *et al.* Indian diabetic retinopathy image dataset (IDRiD): a database for diabetic retinopathy screening research. *Data* **3**, 25 (2018).
21. Kovalyk, O., *et al.* PAPILA: Dataset with fundus images and clinical data of both eyes of the same patient for glaucoma assessment. *Scientific Data* **9**, 291 (2022).
22. Zhang, W., *et al.* Angiographic Report Generation for the 3rd APTOS's Competition: Dataset and Baseline Methods. *medRxiv*, 2023-2011 (2023).
23. Retina Image Bank, available at https://imagebank.asrs.org/.
24. Sabanayagam, C., *et al.* Incidence and progression of diabetic retinopathy: a systematic review. *Lancet Diabetes Endocrinol* **7**, 140-149 (2019).
25. Stein, J.D., Khawaja, A.P. & Weizer, J.S. Glaucoma in Adults-Screening, Diagnosis, and Management: A Review. *JAMA* **325**, 164-174 (2021).
26. Shi, D., He, S., Yang, J., Zheng, Y. & He, M. One-shot Retinal Artery and Vein Segmentation via Cross-modality Pretraining. *Ophthalmol Sci* **4**, 100363 (2024).
27. Wu, B., *et al.* Visual transformers: Token-based image representation and processing for computer vision. *arXiv preprint arXiv:2006.03677* (2020).
28. Liu, X., *et al.* Toward the unification of generative and discriminative visual foundation model: a survey. *The Visual Computer* (2024).
29. Chen, R., *et al.* Translating color fundus photography to indocyanine green angiography using deep-learning for age-related macular degeneration screening. *npj Digital Medicine* **7**, 34 (2024).
30. Song, F., Zhang, W., Zheng, Y., Shi, D. & He, M. A deep learning model for generating fundus autofluorescence images from color fundus photography. *Adv Ophthalmol Pract Res* **3**, 192-198 (2023).
31. Hu, Q., Abramoff, M.D. & Garvin, M.K. Automated separation of binary overlapping trees in low-contrast color retinal images. *Med Image Comput Comput Assist Interv* **16**, 436-443 (2013).
32. Lee, S.S. & Mackey, D.A. Glaucoma - risk factors and current challenges in the diagnosis of a leading cause of visual impairment. *Maturitas* **163**, 15-22 (2022).

33. Tanna, P., Strauss, R.W., Fujinami, K. & Michaelides, M. Stargardt disease: clinical features, molecular genetics, animal models and therapeutic options. *British Journal of Ophthalmology* **101**, 25-30 (2017).
34. Dimaras, H*., et al.* Retinoblastoma. *The Lancet* **379**, 1436-1446 (2012).
35. Li, F., Hu, Z., Chen, W. & Kak, A. Adaptive supervised patchnce loss for learning h&e-to-ihc stain translation with inconsistent groundtruth image pairs. in *International Conference on Medical Image Computing and Computer-Assisted Intervention* 632-641 (Springer, 2023).
36. Unterthiner, T*., et al.* FVD: A new metric for video generation. (2019).
37. Wang, Z., Bovik, A.C., Sheikh, H.R. & Simoncelli, E.P. Image quality assessment: from error visibility to structural similarity. *IEEE transactions on image processing* **13**, 600-612 (2004).
38. Huynh-Thu, Q. & Ghanbari, M. Scope of validity of PSNR in image/video quality assessment. *Electronics letters* **44**, 800-801 (2008).
39. Zhang, R., Isola, P., Efros, A.A., Shechtman, E. & Wang, O. The unreasonable effectiveness of deep features as a perceptual metric. in *Proceedings of the IEEE conference on computer vision and pattern recognition* 586-595 (2018).

## Supplementary Figures

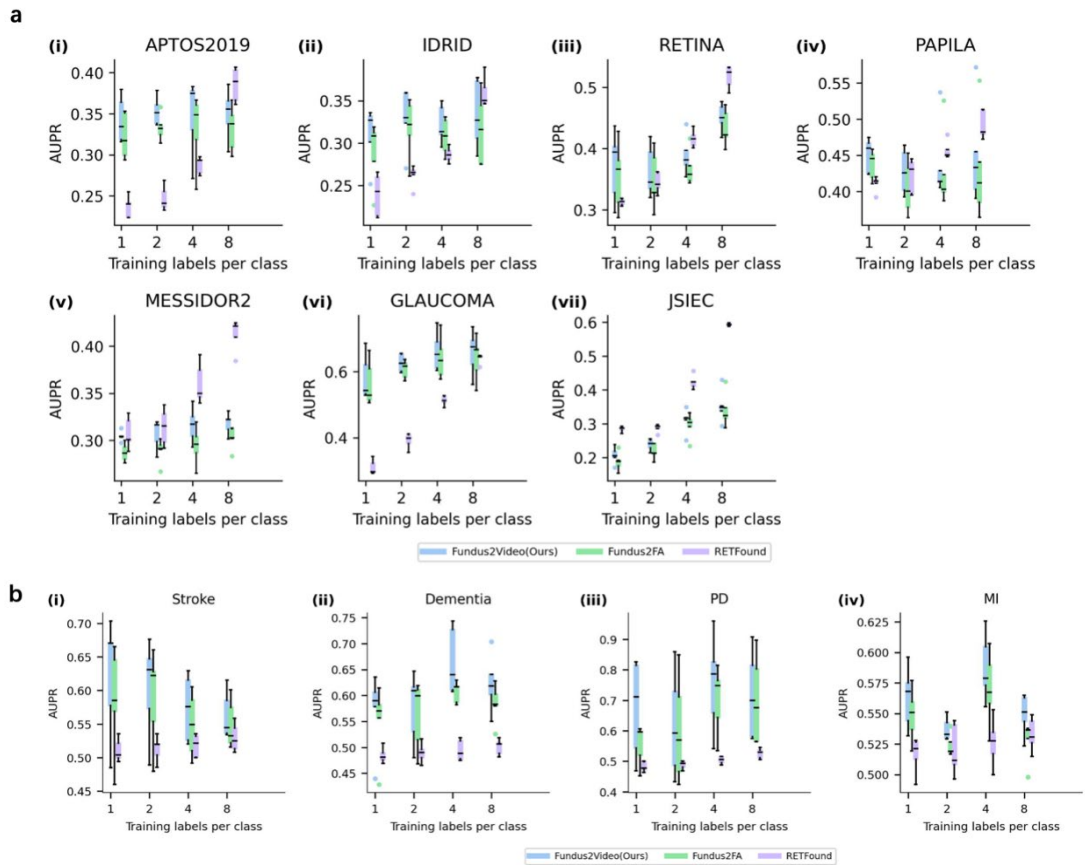

**Supplementary Fig. 1:** Few-shot classification results. AUPR = area under the precision-recall curve. We compared the few-shot learning capabilities of different models across various datasets, considering multiple support numbers for each class. For each support number, we sampled five different sets of training examples and conducted five times of experiments. The boxes represent the quartiles, and the whiskers extend to data points within 1.5 times the interquartile range.